\documentclass[runningheads]{llncs}
\usepackage{graphicx}
\usepackage{csquotes}
\usepackage{subcaption}
\usepackage{booktabs}

\begin{document}
\title{Enhancing Augmentative and Alternative Communication with Card Prediction and Colourful Semantics}
\titlerunning{Enhancing AAC with Colourful Semantics}
%
\author{Jayr Pereira\inst{1,2}\orcidID{0000-0001-5478-438X} \and
Francisco Rodrigues\inst{2} \and
Jaylton Pereira\inst{2} \and
Cleber Zanchettin\inst{2}\orcidID{0000-0001-6421-9747}
Robson Fidalgo\inst{2}\orcidID{0000-0002-4714-2933}}
\authorrunning{J. Pereira et al.}
\institute{
Universidade Federal do Cariri (UFCA), Juazeiro do Norte-CE, Brazil
\email{jayr.pereira@ufca.edu.br}
\and
Centro de Informática, Universidade Federal de Pernambuco (CIn-UFPE), Recife-PE, Brazil
\email{<fasr,jap3,cz,rdnf>@cin.ufpe.br}
}

\maketitle              
\begin{abstract}
This paper presents an approach to enhancing Augmentative and Alternative Communication (AAC) systems by integrating Colourful Semantics (CS) with transformer-based language models specifically tailored for Brazilian Portuguese. We introduce an adapted BERT model, BERTptCS, which incorporates the CS framework for improved prediction of communication cards. The primary aim is to enhance the accuracy and contextual relevance of communication card predictions, which are essential in AAC systems for individuals with complex communication needs (CCN). We compared BERTptCS with a baseline model, BERTptAAC, which lacks CS integration. Our results demonstrate that BERTptCS significantly outperforms BERTptAAC in various metrics, including top-k accuracy, Mean Reciprocal Rank (MRR), and Entropy@K. Integrating CS into the language model improves prediction accuracy and offers a more intuitive and contextual understanding of user inputs, facilitating more effective communication.

\end{abstract}
\section{Introduction}

Augmentative and Alternative Communication (AAC) systems are essential tools designed to aid individuals facing natural speech and language expression challenges \cite{beukelman1998augmentative}. These systems are often utilized by people with complex communication needs (CCN), including those with language impairments caused by cerebral palsy, autism, and stroke-induced aphasia. AAC systems encompass a variety of methods, such as gestures, sign language, and pictograms or communication cards, which are crucial in empowering users to communicate their needs, thoughts, and emotions effectively.

Pictogram prediction, a key component of AAC systems, is vital in facilitating communication \cite{franco2018towards}. It involves using technology to predict and suggest appropriate pictograms or communication cards based on the user's input or context. This predictive capability can enhance the speed and efficiency of communication for AAC users \cite{beukelman1998augmentative}. By accurately anticipating the user's intended message, these systems reduce the cognitive and physical effort in message authoring, enabling a smoother and more natural communication process. Recent research \cite{pereira2023praact} has pointed towards using transformer-based language models to enhance pictogram prediction, showcasing their effectiveness in interpreting user input and context for more accurate suggestions. However, while these models excel in predicting suitable pictograms, a challenge remains: users can construct sentences but often need help with structuring or ordering words appropriately.

Colourful Semantics (CS) \cite{bryan1997colourful} emerges as a transformative tool within this domain. Originally developed as a therapeutic approach to support language structure, CS employs color coding to represent different parts of speech, aiding in the construction of sentences. The adaptation of Colourful Semantics in AAC systems, particularly in the context of pictogram prediction, holds promise for further enhancing the message authoring process. By leveraging the intuitive structure provided by CS, AAC systems can offer more contextually appropriate and semantically rich pictogram suggestions, thereby improving the overall communication experience for users.

This paper aims to demonstrate the efficacy of integrating CS into the predictive models of AAC systems, specifically focusing on Brazilian Portuguese. For this purpose, we employed the PrAACT method \cite{pereira2023praact} to adapt and fine-tune a BERT \cite{devlin-etal-2019-bert} model, originally pre-trained for Brazilian Portuguese \cite{souza2020bertimbau} to perform card prediction using the CS structure. Our methodology involved utilizing the AACPTcorpus \cite{pereira2023icalt}, a synthetic corpus tailored for Brazilian Portuguese, to fine-tune two distinct models: one incorporating the CS structure (BERTptCS) and the other without this structure (BERTptAAC).
The results of our experiments indicate that incorporating the CS structure into the fine-tuning process significantly improves the accuracy and relevance of the model’s predictions. Additionally, it was observed that BERTptCS displayed a more uniform distribution of predicted tokens compared to BERTptAAC. These findings suggest that integrating CS enhances language models' performance in AAC and contributes to a more effective and user-friendly communication tool.

\section{Message Authoring in AAC}

Augmentative and Alternative Communication (AAC) systems employ various tools and techniques to aid individuals with complex communication needs (CCN) in expressing themselves. High-tech AAC systems use pictographic images on communication cards as visual aids, meaningfully representing words and expressions in the user's vocabulary. These systems are especially beneficial for non-literate individuals due to age or disability, facilitating communication even for those with limited cognitive abilities or in the early stages of language development. A notable example of such resources is the ARASAAC database \cite{arassac_2019}, which offers an extensive collection of over 30,000 pictograms in AAC systems.

High-tech AAC systems often arrange these pictograms in grid formats tailored to the specific needs and preferences of the user. This can vary from categorical organization to multi-page systems, but the primary goal remains to enable and streamline the selection of cards for constructing sentences. Anyway, such systems must allow and facilitate card selection for sentence construction \cite{franco2018towards}. Among the strategies that can facilitate message authoring in AAC, we can cite the usage of color coding systems and a word or card prediction technique.

\subsection{Color Coding Systems}

\begin{figure*}[t]
    \centering
    \caption{Color coding systems. (a) the Fitzgerald Keys system, which color cards according to their grammatical class (e.g., noun). (b) Colourful Semantics, which color cards according to their role in a sentence.}
    \begin{subfigure}[b]{\textwidth}
        \centering
        \caption{Fitzgerald keys}
        \includegraphics[width=6cm]{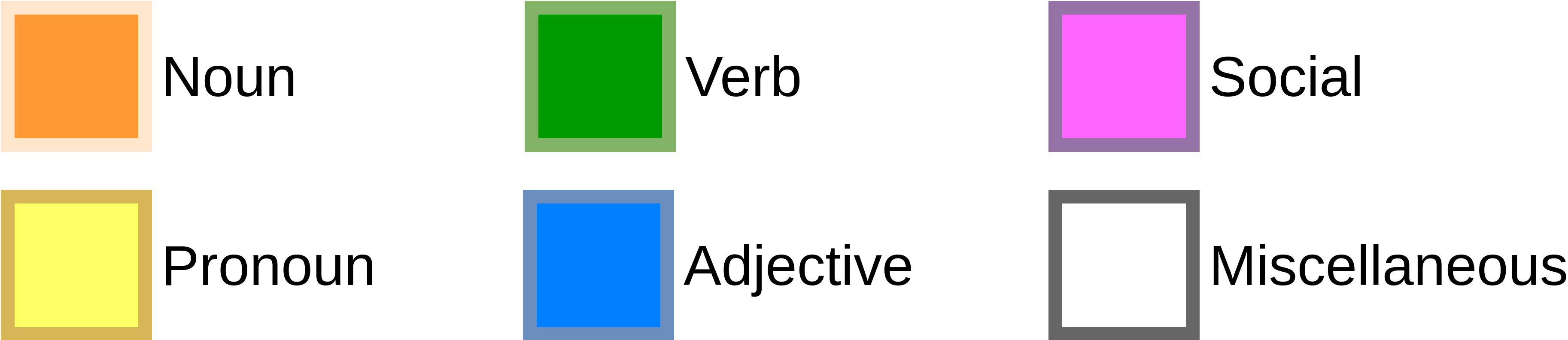}
        
        \label{fig:fitzgeraldkeys}
    \end{subfigure}
    \par\bigskip 
    \begin{subfigure}[b]{\textwidth}
        \centering
        \caption{Colourful semantics}
        \includegraphics[width=10cm]{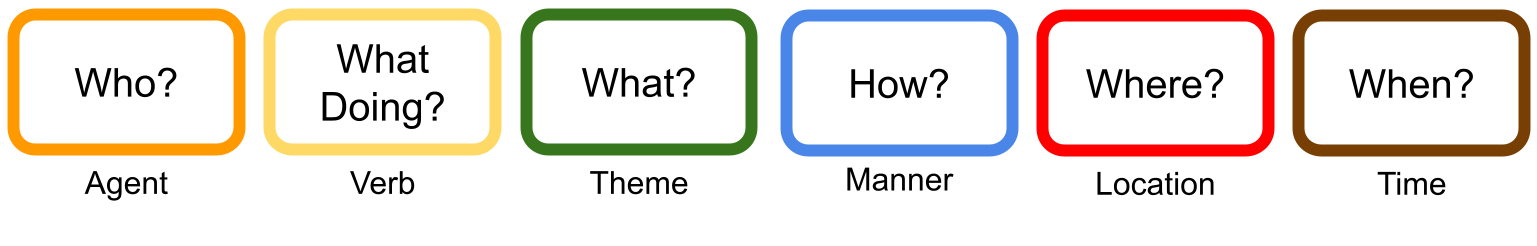}
        
        \label{fig:colourfulsemantics}
    \end{subfigure}
    \label{fig:color_coding}
\end{figure*}

Pictogram selection in robust AAC systems may include color coding and motor planning protocols. A widely adopted system is a modified Fitzgerald Key \cite{fitzgerald1949straight}, categorizing pictograms into six colors based on their grammatical role, as shown in Figure \ref{fig:fitzgeraldkeys}. This method uses colors and suggests a left-to-right organization of pictograms in accordance with their grammatical function.

Colourful Semantics (CS) \cite{bryan1997colourful} is another color coding system, initially developed to assist children with CCN in building and understanding sentences. This system employs colors to delineate semantic roles within a sentence, supplemented by key questions (e.g., Who?) that aid in comprehending the semantic function of each phrase component. CS is distinct in its focus on semantic roles rather than syntactic functions, making it particularly beneficial for individuals with language difficulties. For instance, in a sentence like \textquote{The boy ate popcorn,} CS would identify \textquote{boy} as the Agent and \textquote{popcorn} as the Theme, using different colors and structured questions to reinforce these roles.

\subsection{Card Prediction}

Card prediction is a pivotal aspect of enhancing message authoring in AAC systems, as it involves suggesting relevant communication cards to users during sentence composition. This streamlines the communication process and significantly reduces the cognitive load on users \cite{beukelman1998augmentative}. Recent literature has identified various techniques employed for communication card prediction in AAC systems, with knowledge-based methods being prominently used. For example, semantic grammars and concept networks have been implemented in various studies to enhance prediction accuracy.

Deep learning models have also emerged as potent tools in this domain. Studies by Dudy and Bedrick \cite{Dudy2018} and Pereira et al. \cite{PEREIRA2022pictobert} have utilized neural networks trained with synthetic text corpora. These approaches have shown promise in outperforming traditional statistical models. However, they also pose challenges in terms of computational resources, making their implementation in production settings more complex.
The Predictive Augmentative and Alternative Communication with Transformers (PrAACT) method \cite{pereira2023praact} involves adapting large language models (LLMs) like BERT and GPT for communication card prediction in AAC systems. This method comprises three main steps: Corpus Annotation, Model Fine-Tuning, and Vocabulary Encoding. PrAACT aims to facilitate message authoring in AAC systems through transfer learning by customizing language models to fit the user's vocabulary.


\section{Method}

\begin{figure}[t]
    \centering
    \caption{PrAACT method adapted for adjusting BERTimbau to perform card prediction using Colourful Semantics.}
    \includegraphics[width=10cm]{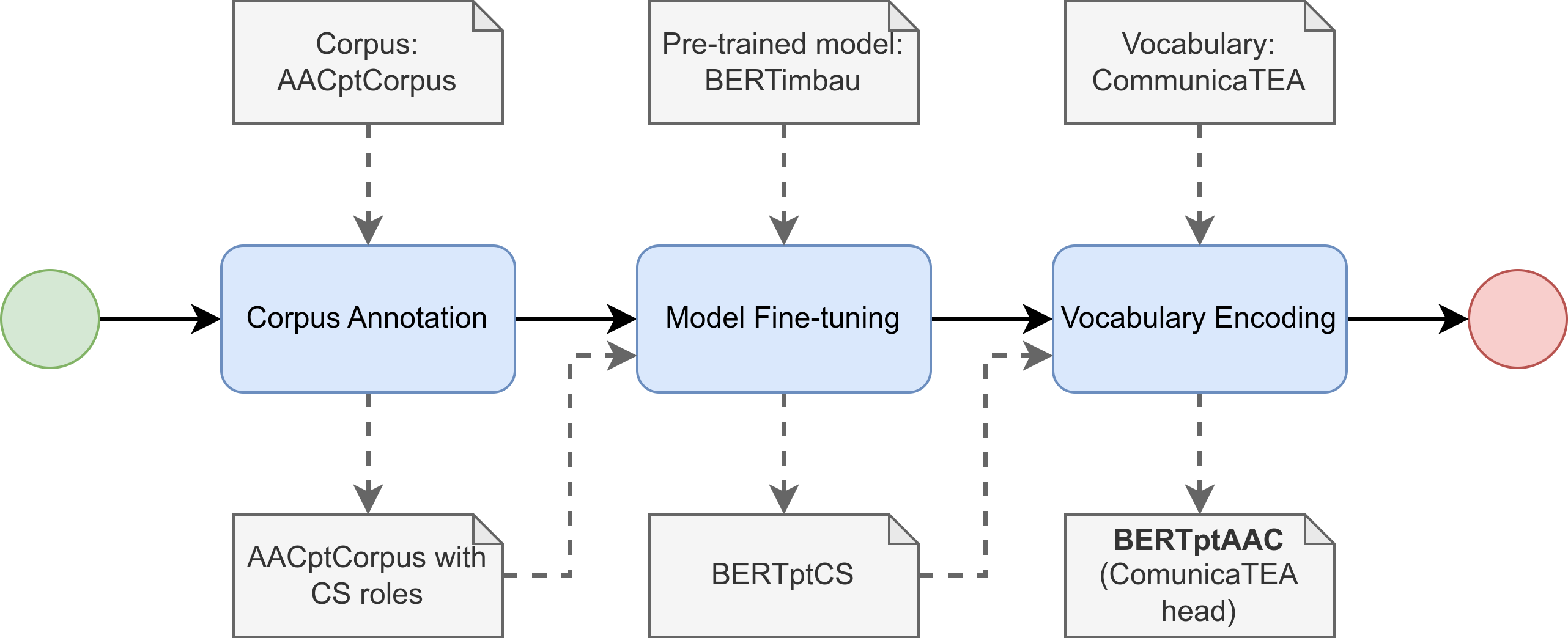}
    \label{fig:BERTptCS}
\end{figure}

We employed the PrAACT method \cite{pereira2023praact} to adapt and fine-tune BERTimbau \cite{souza2020bertimbau}, a BERT model pre-trained for Brazilian Portuguese, to perform card prediction using the CS structure. As shown in Figure \ref{fig:BERTptCS}, the PrAACT method comprises three main steps: Corpus Annotation, Model Fine-Tuning, and Vocabulary Encoding. The following sections detail the methodology employed for each step.

\subsection{Corpus Annotation}

As shown in Figure \ref{fig:BERTptCS}, the first step of the PrAACT method involves annotating the corpus. This step is crucial in customizing the language model to fit the user's vocabulary and communication needs. The AACptCorpus \cite{pereira2023icalt}, a synthetic corpus tailored for Brazilian Portuguese, was used for this purpose. The AACptCorpus comprises 13,000 synthetic sentences used as training data and 667 human-composed sentences used as test data. The sentences in the AACptCorpus were annotated with grammatical structures of the sentences to create semantic scripts like CS. 

For the annotation of ACCptCorpus, the Portuguese model from the Stanza NLP toolkit was employed. This model facilitated the transformation of the corpus into a telegraphic format, simplifying the analysis and extraction of essential information. Additionally, CS roles were applied to achieve a more comprehensive corpus annotation. Syntactic structure and semantic roles were extracted from the corpus through dependency parsing and Semantic Role Labeling (SRL). The dependency parsing, particularly utilizing the Portuguese model from Stanza NLP \cite{qi2020stanza}, was instrumental in delineating the subject+verb+object framework within each sentence.

SRL was used for identifying and categorizing adverbial complements, such as those indicating location, time, and manner. This technique was crucial for pinpointing and designating the semantic roles associated with each component of the sentences. For SRL, the InVeRo semantic parser was utilized \cite{conia-etal-2020-invero}. The synergy of these two methodologies enabled the thorough annotation of our corpus with both syntactic and semantic layers, which were integrated into the CS roles (e.g., who, and what doing).

To illustrate, consider the sentence \textquote{Eu comi pipoca na escola hoje} (I ate popcorn at school today) from AACptCorpus. The subject \textquote{eu} (I), the verb \textquote{comi} (ate), and the object \textquote{pipoca} (popcorn) were discerned through dependency parsing. The adverbial phrase \textquote{na escola} (at school) and the temporal reference \textquote{hoje} (today) were identified by SRL as location and time semantic roles, respectively. Consequently, this facilitated the assignment of semantic roles such as \textquote{who} (eu), \textquote{what doing} (comi), \textquote{what} (pipoca), \textquote{where} (na escola), and \textquote{when} (hoje) to the respective elements of the sentence, following the CS framework.
The resulting annotated corpus contains sentences such as \textquote{$<$quem$>$ eu $</$quem$>$ $<$verbo$>$ querer comer $</$verbo$>$ $<$o\_que$>$ pipoca $</$o\_que$>$}. Using tags to represent the CS roles is important to facilitate the model's understanding of the sentence structure. 

\subsection{Model Fine-tuning}


The similarity between communication card prediction and word prediction tasks is particularly notable in the context of multi-word expressions (MWEs), frequently used in communication cards to succinctly convey complex meanings. For example, single pictograms often represent expressions like \textquote{Good morning}. We added tokens representing MWEs to its existing vocabulary to accommodate these MWEs in BERT. This approach allows BERT to capture the meaning of MWEs more accurately and prevent the loss of semantic information when tokenized into their constituent parts.
For incorporating MWEs into BERT's vocabulary, we extracted two or three-word expressions from the ARASAAAC \cite{arassac_2019} vocabulary for Brazilian Portuguese. We computed the mean vector representation of each MWE from BERT's original input embeddings and added these MWE tokens to the model vocabulary. This method enables BERT to handle MWEs and out-of-vocabulary words using its WordPiece tokenization algorithm. 

We employed BERTimbau \cite{souza2020bertimbau} as the pre-trained model as input for the fine-tuning step. To incorporate the CS roles, such as \textquote{$<$quem$>$} (who) and \textquote{$<$o\_que$>$} (what), into the original vocabulary of the model, we added corresponding vectors to the embedding layers that represent these new tokens. To achieve this, we first tokenized the CS roles and then captured the representation of each token from the BERTimbau original embedding layer. The role representation was obtained as the mean vector of its constituent tokens. For instance, the role \textquote{$<$o\_que$>$} was tokenized as \textquote{$<$}, \textquote{o}, \textquote{\_}, \textquote{que}, \textquote{$>$}, and its corresponding representation vector was computed as the mean vector of these tokens. The reason for using the vector representations of tokens like \textquote{$<$} and \textquote{$>$} is to prevent the model from generalizing the role's meaning to similar tokens. If we just used the representations of \textquote{o} and \textquote{que} tokens, the model might generalize the role \textquote{$<$o\_que$>$} to other contexts where \textquote{o} and \textquote{que} are used together, even when they do not represent the same role. By adding the special tokens \textquote{$<$} and \textquote{$>$}, we ensure that the model learns to associate these representations only with their intended roles, avoiding potential confusion.

Then, we fine-tuned BERTImbau using the AACptCorpus annotated with CS roles. We fine-tuned using a batch size of 384 sequences with 33 tokens (384 * 33 = 12,672 tokens/batch). Each data batch was collated to choose 15\% of the tokens for prediction, following the same rules as BERT: If the $i$-th token is chosen, it is replaced with 1) the $[MASK]$ token 80\% of the time, 2) a random token 10\% of the time or 3) the unchanged $i$-th token 10\% of the time. We use the same optimizer as BERT \cite{devlin-etal-2019-bert}: Adam, with a learning rate of $1 \times 10^{-5}$ for all model versions, with $\beta_1 = 0.9$, $\beta_2 = 0.999$, L2 weight decay of 0.01, and linear decay of learning rate. Fine-tuning was performed in a single 16GB NVIDIA Tesla T4 GPU for 50 epochs. We named the resulting model BERTptCS, where \textquote{pt} refers to the Portuguese language and \textquote{CS} to the Colourful Semantics.

\subsection{Vocabulary Encoding} \label{sec:vocabulary_encoding}

Transformer-based models such as BERT share weights between their input embeddings layer and the decoder layer in the MLM head, a pivotal feature in this encoding process. The decoder layer in BERT predicts masked tokens based on the context (i.e., the entire sequence), receiving word representations from the encoder layers' hidden states. It produces logits, which are converted into probabilities by a softmax function. For Vocabulary Encoding, we modify the decoder layer to produce probabilities for a user-specific vocabulary by updating the weight matrices.

The user vocabulary items are encoded using the model’s original embedding layer, generating embedding vectors for each item. These vectors replace the weights of the original decoder layer, allowing the model to output logits for the new vocabulary. We also incorporate MWEs into the model's vocabulary, addressing regional variations and user preferences. We combine the vectors of their constituent words for MWEs not present in the original vocabulary. For instance, if the expression \textquote{fazer xixi} is not in the model vocabulary, its communication card vector is derived from the average of the vectors for \textquote{fazer} and \textquote{xixi}. This process generates a $h \times |V|$ matrix, where $h$ represents the model's hidden states size and $|V|$ is the size of the user vocabulary. This matrix replaces the decoder layer's weights, enabling it to produce logits corresponding to the new vocabulary.

As input for this step, we used the communication cards from a vocabulary constructed by AAC specialists from ComunicaTEA\footnote{\url{https://comunicatea.com.br/}}, a Brazilian association formed by parents of children with CCN aimed at helping other families to have access to and use AAC tools. The vocabulary was constructed using the Reaact AAC Platform\footnote{\url{https://reaact.com.br/}} and is freely available for testing\footnote{Access \url{https://login.reaact.com.br/login} and click at \textquote{Testar Flipbook}.}. 
Each communication card has a picture (pictogram or photo) and a caption with the word or expression represented by the picture. The first screen presents the most frequently used cards to the user, along with folders marked with dashed border lines such as \textquote{pessoas}, \textquote{ações}, \textquote{comidas e bebidas}, \textquote{lugares} and \textquote{animais}. 
    

The encoding process for BERTptCS followed the following steps:
1) we used the BERTimbau tokenizer to tokenize the user vocabulary items, generating a list of tokens;
2) we extracted the embedding vectors for each token from the BERTimbau's input embeddings layer;
3) we calculated the mean of these vectors, generating a single vector for each vocabulary item;
This ensures that the embeddings comprehensively represent the communication card. Subsequently, the BERTimbau MLM head decoder layer weights were replaced with these embeddings. This modification enables the model to produce a probability distribution over the ComunicaTEA vocabulary for sentence completions, enhancing the system's communication card prediction capabilities.

\section{Experiments}

We conducted experiments to evaluate the performance of the BERT model fine-tuned with the CS structure (BERTptCS) and a model fine-tuned without this structure, which we named BERTptAAC. We performed a completion evaluation experiment to assess the models' quality in predicting appropriate communication cards to complete a given sentence. Through this experiment, we aimed to identify the best-performing model and potential areas for further improvement in the prediction of communication cards. The goal was to assess whether incorporating CS roles improves the prediction accuracy of communication cards.

For fine-tuning BERTptAAC, the process is similar to that of BERTptCS. However, in BERTptACC, we do not use CS tags (e.g., $<$quem$>$ $</$quem$>$) in the training text. Instead, we keep the words in the corresponding roles positions: $<$who$>$ $<$verb$>$ $<$what$>$ $<$how$>$ $<$where$>$ $<$when$>$ to ensure the usage of direct order sentences. Unlike BERTptCS, there is no modification in the model's input vocabulary for BERTptACC. Additionally, the training dataset for BERTptAAC contains more minor sequences than BERTptCS, requiring fewer computational resources.

For this experiment, we used the test set from the AACptCorpus, consisting of 667 sentences constructed by humans. To evaluate the ability of the models to predict appropriate communication cards to complete a given sentence, we masked the last token of each sentence, excluding the CS roles tags. To illustrate, a sentence like \textquote{$<$quem$>$ eu $<$/quem$>$ $<$verbo$>$ comer $<$/verbo$>$ $<$/o\_que$>$ pipoca $<$/o\_que$>$} is transformed into \textquote{$<$quem$>$ eu $<$/quem$>$ $<$verbo$>$ comer $<$/verbo$>$ $<$/o\_que$>$ [MASK] $<$/o\_que$>$}. For the BERTptAAC model, we mask the sentence as follows: \textquote{eu comer [MASK]}.

Previous work on communication card prediction has used keystroke saving, Mean Reciprocal Rank (MRR), top-k accuracy, and perplexity as metrics for automatically evaluating pictogram prediction models \cite{Dudy2018,PEREIRA2022pictobert}. However, perplexity may not be a suitable metric for assessing models' quality in a sentence completion task. Perplexity is commonly used to evaluate language models based on their ability to predict the next word given some context. However, in a sentence completion task where the model is asked to predict the missing word in a given sentence, the focus is not only on the likelihood of the predicted word but also on whether the correct word makes sense in the context. Still, it does not consider the appropriateness or relevance of the predicted items. Additionally, keystroke saving is unsuitable for our experiment, as the methods we are comparing will not change the AAC system grid or folders, so the number of selections to construct a given sentence should be the same. Therefore, in this experiment, we use top-k accuracy (ACC@K) and MRR to assess the quality of the models' predictions. 
For ACC@K, we use different K values (1, 9, 18, 25, 36) to simulate the grid sizes in AAC systems. ACC@K measures the proportion of times the expected communication card appears within the top K predicted cards.

In addition to top-k accuracy and MRR, we add Entropy@K to our metrics set, as it provides insight into the diversity of the predicted pictograms, which is essential in AAC scenarios where users may have a limited vocabulary. Entropy@K measures the uncertainty of the top-K items suggested by a model. The metric calculates the entropy of the probability distribution of the top-K predictions. This way, the higher the Entropy@K score, the more uncertain the model's predictions are. Entropy@K can be calculated as:
\( Entropy@K = -\frac{1}{K} \sum_{i=1}^K \log(p(y_i|X)) \),
where $K$ is the number of predictions to consider, $p(y_i|X)$ is the predicted probability (in log scale) of the $i$-th pictogram given the input sentence $X$. This equation measures the entropy or \textquote{surprise} of the model's predictions up to the top $K$ items for each input. A lower score indicates more confident and consistent predictions across different inputs.

\section{Results}

Table \ref{tab:results_AACatK_MRR} shows the results of the top-n accuracy (ACC@K) and Mean Reciprocal Rank (MRR) of BERTptCS and BERTptAAC. The results demonstrate that both models, BERTptCS and BERTptAAC, perform better at higher K values, as indicated by the increasing values of ACC@K. Additionally, BERTptCS outperforms BERTptAAC in all ACC@K metrics, with the most significant difference being at ACC@1. However, when looking at MRR, we can see that BERTptCS also outperforms BERTptAAC but with a smaller margin. Yet, on average, BERTptCS is better at ranking the correct communication card prediction in the top positions of the list of candidates compared to BERTptAAC. 
Overall, these results suggest that using the CS structure can improve the accuracy of the communication card prediction model.

\begin{table}[]
\centering
\caption{Results of top-n accuracy (ACC@K) and Mean Reciprocal Rank (MRR).}

\footnotesize
\begin{tabular}{@{}lcccccc@{}}
\toprule
Model     & ACC@1 & ACC@9 & ACC@18 & ACC@25 & ACC@36 & MRR  \\ \midrule
BERTptCS  & 0,50  & 0,74  & 0,81   & 0,85   & 0,87   & 0,58 \\
BERTptAAC & 0,45  & 0,69  & 0,77   & 0,80   & 0,84   & 0,53 \\ \bottomrule
\end{tabular}
\label{tab:results_AACatK_MRR}
\end{table}

Table \ref{tab:results_EntropyAtK} presents the results of Entropy@K for the BERTptCS and BERTptAAC models. The entropy measures the uncertainty of the distribution of the predicted communication cards. The lower the entropy, the more confident the model is in its predictions. The table shows that both models have higher entropy scores as the value of K increases. BERTptCS has lower entropy scores than BERTptAAC for all values of K, which indicates that BERTptCS is better at predicting the correct communication card, as it has a more concentrated distribution of probabilities. Additionally, BERTptCS has an entropy of 0.31 for Entropy@1, indicating that it makes a very confident prediction for the top candidate. In contrast, BERTptAAC has an entropy of 1.02, indicating it is less confident in its top prediction. This suggests that incorporating the CS structure in the training process can lead to more accurate and confident predictions.

\begin{table}[b]
\centering
\caption{Results of Entropy@K for BERTptCS and BERTptAAC.}
\footnotesize
\begin{tabular}{@{}lccccc@{}}
\toprule
Model     & Entropy@1 & Entropy@9 & Entropy@18 & Entropy@25 & Entropy@36 \\ \midrule
BERTptCS  & 0,31      & 19,73     & 36,06      & 47,74      & 79,91      \\
BERTptAAC & 1,02      & 22,68     & 45,57      & 59,42      & 95,43      \\ \bottomrule
\end{tabular}
\label{tab:results_EntropyAtK}
\end{table}

The results of the experiments show that BERTptCS outperforms BERTptAAC in all of the metrics evaluated. Specifically, the ACC@K metric shows that BERTptCS has higher accuracy than BERTptAAC for all K values. Furthermore, the MRR values of BERTptCS are also higher, indicating that BERTptCS provides more relevant and accurate predictions. In addition, the Entropy@K metric shows that BERTptCS produces more uniform distributions of predicted tokens across all positions than BERTptAAC. These results demonstrate that incorporating the CS structure into the fine-tuning process of BERT improves the accuracy and relevance of the model's predictions and the uniformity of the predicted tokens' distribution. Therefore, we conclude that BERTptCS is a better model than BERTptAAC for predicting communication cards. 

\subsection{Models' Predictions Analysis} \label{sec:ModelAnalysisPortuguese}

This section analyzes the two Brazilian Portuguese models, BERTptCS and BERTptAAC. The analysis focuses on examining the output of the models and identifying patterns and trends in the predictions. Through a qualitative analysis, we aim to understand better how each model performs regarding predicted tokens' accuracy, relevance, and distribution. This analysis provides insights into the strengths and limitations of each model, which can be used to guide future development and improvement of AAC systems. The analysis is conducted considering the communication cards present in the ComunicaTEA vocabulary.

\begin{figure}[t]
    \caption{Models' predictions for the beginning of sentence construction.}
    \centering
    \begin{subfigure}[b]{\textwidth}
        \centering
        \caption{BERTptCS top-12 predictions for \textquote{$<$quem$>$ [MASK] $</$quem$>$}.}
        \includegraphics[width=\textwidth]{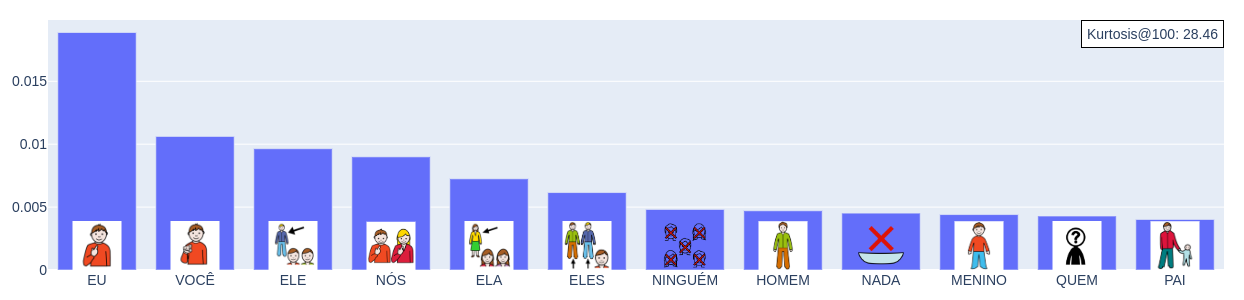}
        
    \end{subfigure}
    \begin{subfigure}[b]{\textwidth}
        \centering
        \caption{BERTptAAC top-12 predictions for \textquote{[MASK]}.}
        \includegraphics[width=\textwidth]{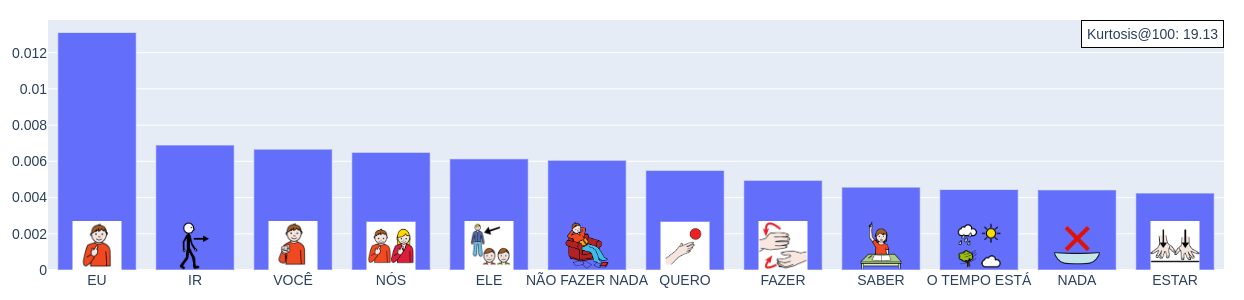}
        
    \end{subfigure}

    \label{fig:subject_prediction}
\end{figure}

Figure \ref{fig:subject_prediction} shows examples of predictions performed by BERTptCS and BERTptAAC, simulating the beginning of sentence construction. The communication cards in the figure are composed of the pictogram images, the caption, and the probability the model has given to each card. Analyzing the figure, we notice that the BERTptCS predictions consist mainly of agentive words, which can play the role of the agent in a sentence (e.g., \textquote{eu}, i.e., \textquote{I}). The predictions made by BERTptAAC are a mix of verbs and pronouns, as the model has no information on how the user wants to initiate the construction of the sentence, either by the subject or by the verb. The use of CS roles brings more context to the prediction, and this can be verified in the example shown, which simulates the user choosing the first sentence communication card. Notice that if the user prefers to start from the verb, they can do so, and the model will predict the most suitable communication card to fill the mask in \textquote{$<$verbo$>$ [MASK] $</$verbo$>$}. Examples like this highlight the importance of using CS roles, as they can treat the sentence components differently, providing more accurate predictions.

The Kurtosis@100 values for both BERTptCS and BERTptAAC at the beginning of sentence construction are quite high, which may indicate that both models are overfitting to the training data and cannot generalize well to new examples. The small number of training examples could cause this overfitting, but it could also be due to the characteristics of the training corpus used for fine-tuning. The corpus has many sentences beginning with the word \textquote{eu}, the most frequent word. However, it is worth noting that the human-composed sentences, which AAC specialists informed as the ones they consider common in AAC, also have many sentences starting with \textquote{eu} \cite{pereira2023icalt}. This suggests that this may be a characteristic of the AAC domain.

\begin{figure}[h!]
    \caption{Models' predictions for verb completion.}
    \centering
    \begin{subfigure}[b]{\textwidth}
        \centering
        \caption{BERTptAAC top-12 predictions for \textquote{eu quero comer [MASK]}.}
        \includegraphics[width=\textwidth]{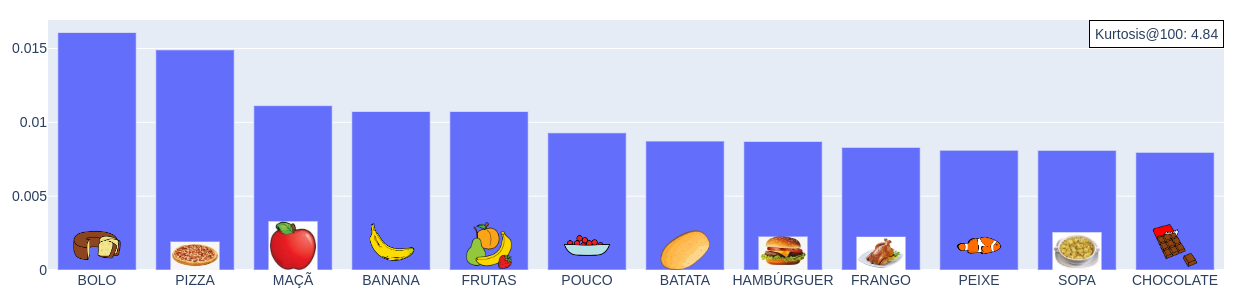}
        
    \end{subfigure}
    \begin{subfigure}[b]{\textwidth}
        \centering
        \caption{BERTptCS top-12 predictions for \textquote{$<$quem$>$ eu $</$quem$>$ $<$verbo$>$ quero comer $</$verbo$>$ $<$o\_que$>$ [MASK] $<$/o\_que$>$}.}
        \includegraphics[width=\textwidth]{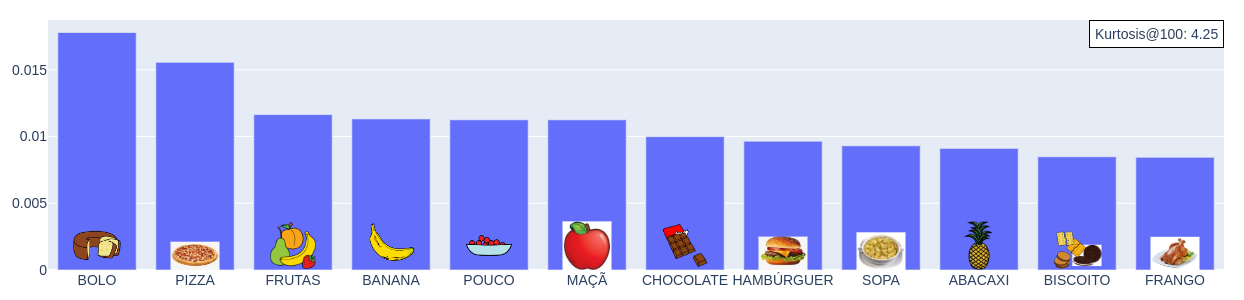}

    \end{subfigure}
    \begin{subfigure}[b]{\textwidth}
        \centering
        \caption{BERTptCS top-12 predictions for \textquote{$<$quem$>$ eu $</$quem$>$ $<$verbo$>$ quero comer $</$verbo$>$ $<$onde$>$ [MASK] $<$/onde$>$}.}
        \includegraphics[width=\textwidth]{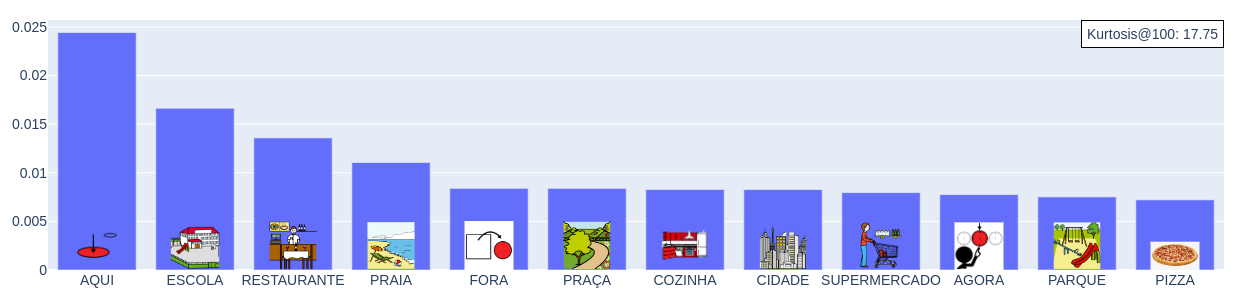}
        
        \label{fig:location_complement}
    \end{subfigure}
    \begin{subfigure}[b]{\textwidth}
        \centering
        \caption{BERTptCS top-12 predictions for \textquote{$<$quem$>$ eu $</$quem$>$ $<$verbo$>$ quero comer $</$verbo$>$ $<$quando$>$ [MASK] $<$/quando$>$}.}
        \includegraphics[width=\textwidth]{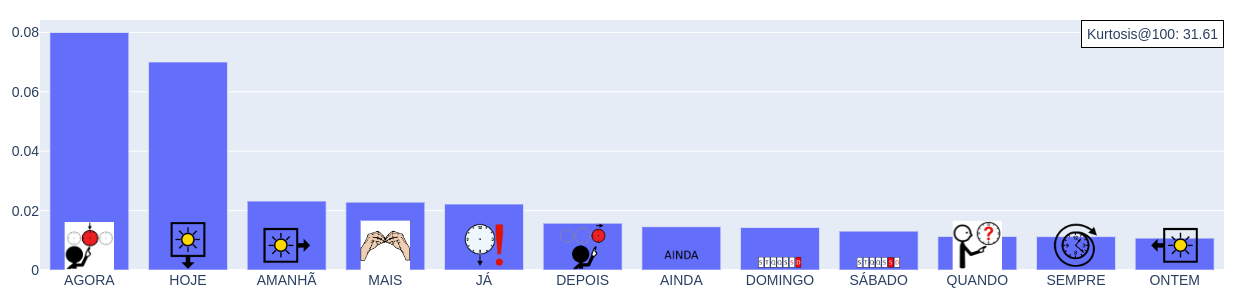}
        
        \label{fig:time_complement}
    \end{subfigure}

    \label{fig:eu_quero_comer}
\end{figure}

In Figure \ref{fig:eu_quero_comer}, we present examples of predictions for inserting the verb object complement in the sentence \textquote{eu quero comer} (I want to eat). The predictions made by BERTptCS and BERTptAAC are quite similar if we consider using a $<$o\_que$>$ mark for the CS model. Both the models, BERTptCS and BERTptAAC, assigned high probabilities to eatable things when predicting the verb object complement for the sentence \textquote{eu quero comer} (I want to eat).
The predictions made by BERTptAAC for inserting the verb object complement in the sentence \textquote{eu quero comer} are affected by the structure of the training corpus.
If we compare BERTptAAC to its equivalent English version in this work, the BERT-AAC few-shot, we notice that both models assigned high probabilities to non-eatable things to complete the sentence \textquote{I want to eat [MASK]}.
BERTptAAC was fine-tuned using a version of AACptCorpus that does not use CS roles but maintains the words in the sentences following the CS order (e.g., who, what doing, what, how, where). Therefore, it is expected that in the training corpus, complements for \textquote{what} occur with high frequencies.

One advantage of BERTptCS over BERTptAAC is its ability to make more accurate predictions using other roles besides the complement. For example, suppose the user wants to construct a sentence indicating where he/she wants to eat. Using BERTptAAC as we trained it would require the user first to inform what they want to eat, or they would see communication cards for locations with a considerably lower probability. However, with CS, the user can provide more context to the model, which is useful for making more accurate predictions, as demonstrated in Figure \ref{fig:eu_quero_comer}. For example, Figure \ref{fig:location_complement} shows the predicted complements for the location's role. All the communication cards suggest locations except for the card with \textquote{Pizza}. 
Figure \ref{fig:time_complement} shows the top-12 predicted complements for the time role. All the predicted complements for time are consistent, except for \textquote{mais} (more), which must have occurred together with time complements during training (e.g., \textquote{mais tarde}, i.e., \textquote{later}). It is important to note that the annotation process can sometimes group unrelated words, leading to unexpected predictions. Also, slight differences in the training data can lead to differences in the model's predictions.



\section{Conclusions}

This study investigates the integration of Colourful Semantics (CS) into Augmentative and Alternative Communication (AAC) systems, specifically focusing on enhancing communication card prediction using transformer-based language models for Brazilian Portuguese. Our approach involved adapting and fine-tuning a BERT model, BERTptCS, to incorporate CS structures and comparing its performance against a model without CS integration, BERTptAAC. 
Our findings demonstrate that incorporating CS into the language model improves the accuracy and relevance of communication card predictions. The BERTptCS model outperformed the BERTptAAC model across various metrics, including top-k accuracy, Mean Reciprocal Rank (MRR), and Entropy@K. This indicates that the CS structure enhances the model's ability to predict appropriate communication cards.

Furthermore, applying CS within the model facilitated a more intuitive and contextual prediction process. The predictions became more aligned with the users' communicative intentions by enabling the model to consider semantic roles and relationships within sentences. This aspect is particularly crucial in AAC systems, where the goal is to support individuals with complex communication needs in expressing themselves accurately and efficiently.
However, the study also identified challenges and areas for future research. The design of user interfaces that effectively incorporate CS and the need for further optimization to handle the diverse and dynamic nature of language in AAC contexts are areas that warrant additional exploration. Extending this research to other languages and dialects could provide broader insights and benefits.


%
%
%
%
\bibliographystyle{splncs04}

\bibliography{main}

\begin{thebibliography}{10}
\providecommand{\url}[1]{\texttt{#1}}
\providecommand{\urlprefix}{URL }
\providecommand{\doi}[1]{https://doi.org/#1}

\bibitem{beukelman1998augmentative}
Beukelman, D.R., Light, J.C.: Augmentative \& Alternative Communication: Supporting Children and Adults with Complex Communication Needs. Paul H. Brookes Baltimore, fifth edition edn. (2020)

\bibitem{bryan1997colourful}
Bryan, A.: Colourful Semantics: Thematic Role Therapy, chap.~3.2, pp. 143--161. John Wiley \& Sons, Ltd (2003). \doi{10.1002/9780470699157.ch10}

\bibitem{conia-etal-2020-invero}
Conia, S., Brignone, F., Zanfardino, D., Navigli, R.: {I}n{V}e{R}o: Making semantic role labeling accessible with intelligible verbs and roles. In: Proceedings of the 2020 Conference on Empirical Methods in Natural Language Processing: System Demonstrations. pp. 77--84. Association for Computational Linguistics, Online (2020). \doi{10.18653/v1/2020.emnlp-demos.11}

\bibitem{devlin-etal-2019-bert}
Devlin, J., Chang, M.W., Lee, K., Toutanova, K.: {{BERT}: Pre-training of Deep Bidirectional Transformers for Language Understanding}. In: Proceedings of the 2019 Conference of the North {A}merican Chapter of the Association for Computational Linguistics: Human Language Technologies, Volume 1 (Long and Short Papers). pp. 4171--4186. Association for Computational Linguistics, Minneapolis, Minnesota (2019). \doi{10.18653/v1/N19-1423}

\bibitem{Dudy2018}
Dudy, S., Bedrick, S.: {Compositional Language Modeling for Icon-Based Augmentative and Alternative Communication}. Proceedings of the conference. Association for Computational Linguistics. Meeting  \textbf{2018},  25--32 (jul 2018). \doi{10.18653/v1/w18-3404}

\bibitem{fitzgerald1949straight}
Fitzgerald, E.: Straight language for the deaf: a system of instruction for deaf children. Volta Bureau (1949)

\bibitem{franco2018towards}
Franco, N., Silva, E., Lima, R., Fidalgo, R.: Towards a reference architecture for augmentative and alternative communication systems. In: Brazilian Symposium on Computers in Education (Simp{\'o}sio Brasileiro de Inform{\'a}tica na Educa{\c{c}}{\~a}o-SBIE). vol.~29, p.~1073 (2018). \doi{10.5753/cbie.sbie.2018.1073}

\bibitem{arassac_2019}
Palao, S.: Arasaac: Aragonese portal of augmentative and alternative communication (2019), \url{http://www.arasaac.org/}

\bibitem{pereira2023icalt}
Pereira, J., Nogueira, R., Zanchettin, C., Fidalgo, R.: An augmentative and alternative communication synthetic corpus for brazilian portuguese. In: 2023 IEEE International Conference on Advanced Learning Technologies (ICALT). pp. 202--206 (2023). \doi{10.1109/ICALT58122.2023.00066}

\bibitem{PEREIRA2022pictobert}
Pereira, J.A., Macêdo, D., Zanchettin, C., {de Oliveira}, A.L.I., do~Nascimento~Fidalgo, R.: {PictoBERT}: Transformers for next pictogram prediction. Expert Systems with Applications  \textbf{202},  117231 (2022). \doi{https://doi.org/10.1016/j.eswa.2022.117231}

\bibitem{pereira2023praact}
Pereira, J.A., Pereira, J.A., Zanchettin, C., {do Nascimento Fidalgo}, R.: {PrAACT}: Predictive augmentative and alternative communication with transformers. Expert Systems with Applications  \textbf{240},  122417 (2024). \doi{10.1016/j.eswa.2023.122417}

\bibitem{qi2020stanza}
Qi, P., Zhang, Y., Zhang, Y., Bolton, J., Manning, C.D.: Stanza: A {Python} natural language processing toolkit for many human languages. In: Proceedings of the 58th Annual Meeting of the Association for Computational Linguistics: System Demonstrations (2020), \url{https://nlp.stanford.edu/pubs/qi2020stanza.pdf}

\bibitem{souza2020bertimbau}
Souza, F., Nogueira, R., Lotufo, R.: Bertimbau: Pretrained bert models for brazilian portuguese. In: Cerri, R., Prati, R.C. (eds.) Intelligent Systems. pp. 403--417. Springer International Publishing, Cham (2020)

\end{thebibliography}

\end{document}